# Artificial Intelligence for Suicide Assessment using Audiovisual Cues: A Review

Sahraoui Dhelim [1], Liming Chen [2], Huansheng Ning [3*] and Chris Nugent [2]

[1] School of Computer Science, Dublin College University, Ireland
[2] School of Computing, Ulster University, United Kingdom
[3] School of Computer and Communication Engineering, University of Science and Technology Beijing, China
[*] Corresponding author: ninghuansheng@ustb.edu.cn

**Abstract**

Death by suicide is the seventh leading death cause worldwide. The recent advancement in Artificial Intelligence (AI), specifically AI applications in image and voice processing, has created a promising opportunity to revolutionize suicide risk assessment. Subsequently, we have witnessed fast-growing literature of research that applies AI to extract audiovisual non-verbal cues for mental illness assessment. However, the majority of the recent works focus on depression, despite the evident difference between depression symptoms and suicidal behavior non-verbal cues. In this paper, we review the recent works that study suicide ideation and suicide behavior detection through audiovisual feature analysis, mainly suicidal voice/speech acoustic features analysis and suicidal visual cues. Automatic suicide assessment is a promising research direction that is still in the early stages. Accordingly, there is a lack of large datasets that can be used to train machine leaning and deep learning models proven to be effective in other, similar tasks.

**1. Introduction**

Suicide is a global mental health problem [1]. More than 800,000 people pass away from suicide, and more than 16 million people attempt suicide every year [1]. Suicide is currently the fourth leading cause of death for people aged between 15 and 29 years [2]. It is still a challenge to assess suicide risks and detect suicide due to the transient and ambivalent nature of severe suicide intent, and the patient's hesitancy. Most suicide risk assessment methods depend on voluntary disclosure of the patient, and most suicidal patients deny suicide ideation during an interview [3][4]. In such cases, the clinician relies on various secondary information sources, such as the patient's health records, the nurse's observation, and other secondary factors. However, the final judgment about the suicide risk level is based on the clinician's intuition and observation of the patient's behaviors during the interview. Although the clinician can observe obvious emotions and body gestures, the clinician cannot observe micro facial expressions and biosignals such as heartbeat rate and brain waves. Therefore, building an Artificial Intelligence (AI) enabled suicide risk assessment tool that observes the patient's biomarker and audiovisual cues and learns suicide ideation patterns might be a decisive indicator that could help the clinician to reach a clear judgment [5]. With the recent advances in the field of user-centered computing [6], [7], automatic mental disorders detection using audiovisual data has become an active research topic. The patient's acoustic and visual markers are analysed to detect mental disorders such as depression [8]. The recent advancement in AI, specifically AI applications in image, voice processing and natural language processing [9], [10], has created a promising opportunity to revolutionize suicide risk assessment. Subsequently, we have witnessed fast-growing literature of research that applies AI to extract audiovisual non-verbal cues for mental illness assessment [11]. Although that previous research has found associations between acoustic features and suicide tendency [12][13]; suicidal patients have lower acoustic energy, breathy voice quality, and abnormal glottal control compared to people without suicide ideation history [14]. However, the majority of the recent works focus on depression [15], despite the evident difference between depression signs and suicidal behavior non-verbal cues. In this paper, we review the recent works that study suicide ideation and suicide behavior detection through audiovisual feature analysis, mainly suicidal voice/speech acoustic features analysis and suicidal visual cues analysis. Suicidal ideation can vary in presentation and severity, but generally it can be classified as one of the following phases [16]. (1) passive suicide ideation: when the subject has thoughts of suicide or self-harm but no plan to carry it out. (2) active suicide ideation: when the subject has thoughts of engaging in suicide-related behavior and has suicidal intent and/or had developed a plan to carry it out. (3) suicidal attempt: the subject had attempted suicide and/or still attempting to suicide following an unsuccessful suicide attempt [17].

The focus of the current review is automatic suicide ideation and suicide behavior detection through audiovisual feature analysis, with a special focus on works that used machine learning and deep learning approaches. We limit the coverage of the review to works published between 2004 and 2021. We have used PRISMA (Preferred Reporting Items for Systematic Reviews and Meta-Analyses) framework [18] guidelines to select publications related to automatic suicidal behavior detection. As shown in Figure 1, initially, 251 related papers between January 2004 and November 2021 were identified after searching Google Scholar, Elsevier, Web of Science, ACM Digital Library, IEEE Xplore digital library, Springer and PubMed for articles related to the following research queries: "automatic suicide detection", "machine learning suicide detection", "suicidal speech analysis", "suicidal visual cues", "suicide ideation detection". The searches were limited to articles written in English. 455 additional articles were identified as related works; these papers were found by tracking the citation map of the searched articles. After removing duplicated articles, a total of 706 papers were collected in the identification phase. In the screening phase, based on the title and abstract screening 551 papers were excluded for not meeting the inclusion criteria. The majority of these papers either studied suicidal behavior from a clinical perspective, without automatic detection, or they use automatic detection for detecting mental disorders such as anxiety and minor depression. 104 articles were excluded in the eligibility phase after full-text reading. Finally, 51 articles were qualified for final inclusion.

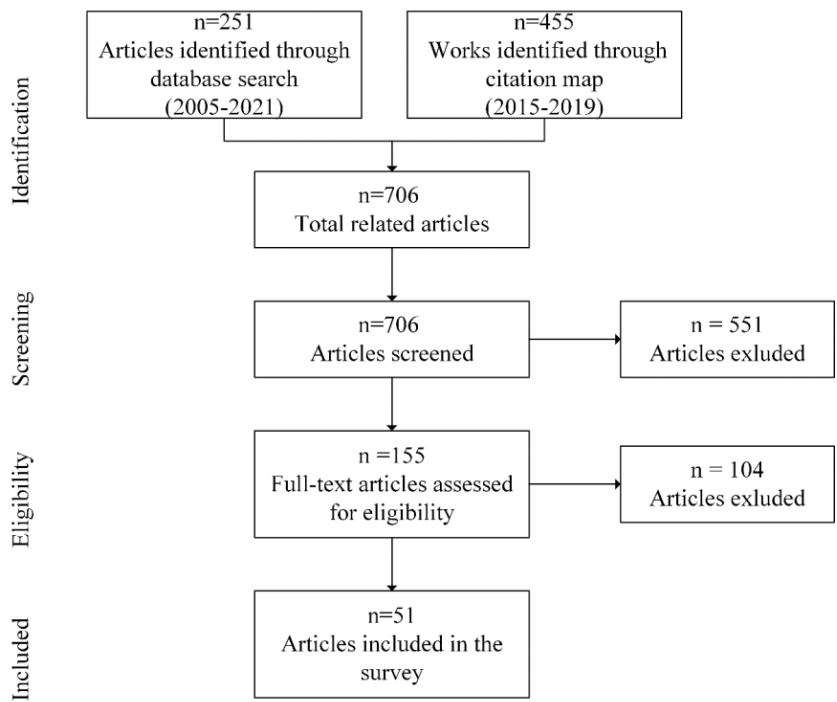

**Figure 1 PRISMA flowchart of the review phases**

The suicidal marker that can be extracted from audiovisual data, can be roughly divided into two main categories. Audio markers that convey the suicidal cue can be obtained by analyzing the acoustic features of the speech, and visual markers that can be observed in the suicidal patient's upper body behaviors during suicidal assessment procedures such as a clinical interview. Figure 2 shows the most used acoustic and visual features used in the literature on suicidal assessment using audiovisual data.

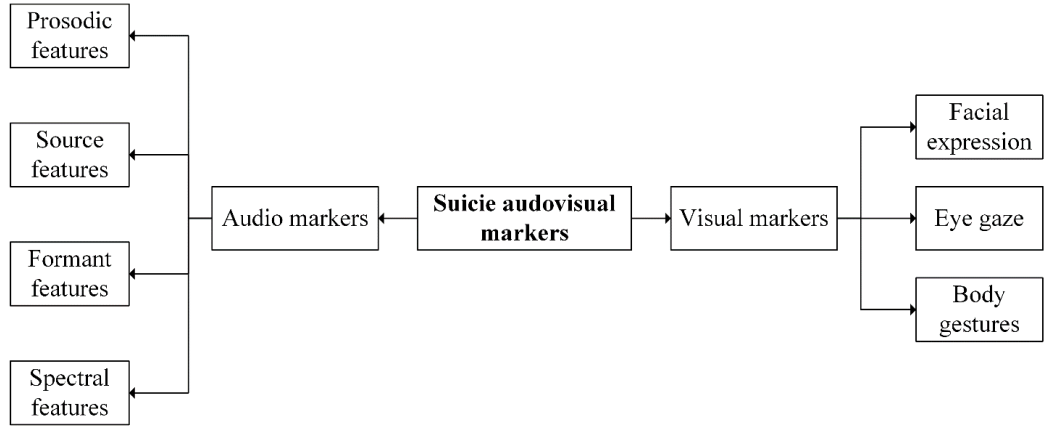

**Figure 2 Suicidal audiovisual markers classification**

The rest of the paper is organized as follows:

In Section 2, we discuss data collection scenarios of the reviewed works and summarize some of the most used audiovisual datasets for automatic suicide assessment. In Section 3, we focus on speech analysis and audio-based suicide assessment. In Section 4, we review the works that used upper body features extracted from visual cues for suicide assessment. While Section 5 focuses on the detection models applied to audiovisual data for suicide behavior detection. In Section 6, we summarize the findings and outline limitations and potential future directions. Finally, Section 7 concludes the review.

## 2. Data collection

Suicide audiovisual data can be collected using three methods. (1) through an open interview between the patient and the clinician; (2) by capturing the audiovisual recording during casual conversations; (3) or by asking patients to read a speech or a set of sentences. Each one of these methods has its advantages and drawbacks. Because the recorded sentence is the same for all participants, the reading task method makes it easy to distinguish a suicidal patient from non-suicidal controls, which enable the machine learning model to focus on the audiovisual feature differences between suicidal patients and the control participants. The second advantage of the reading task method is that the model requires fewer data to learn, as it does not need to learn the context of the spoken sentence from the data, unlike the interview method, where the model needs to deduce the context of the spoken sentences.

In the interview method, unlike the reading task, the reason behind sudden changes in one of the acoustic markers must be distinguished from the context of the interview. For example, is the detected change due to grammatical contexts, such as intonation, or a suicidal cue? The second advantage of the reading task is that the clinician can design the reading passage to target specific sounds or syllables that are related to certain suicidal cue patterns. One of the famous reading tasks is known as the "rainbow passage", which is used in speech science because it contains all of the normal sounds in spoken English, and is phonetically balanced [19]. Table 1 presents study samples and data collection scenarios of some reviewed works.

Whereas Table 2 summarizes some of the most used datasets that contain audiovisual recordings along with ground truth label regarding the mental conditions of the recoded individuals.

**Table 1 Studied samples and data collection scenarios**

| Ref | N | Sample | Collection scenario |
|---|---|---|---|
| [20] | 124 | military couples | Recorded home conversations of veteran couples, following that they subjects were asked to take reasons-for-living and relationship counselling interviews. The interviews are 10 minutes long where the subjects were asked to discuss about life meaning and the their relationship conflicts. |
| [21][22] | 60 | Adolescent | Total of 30 male and 30 females were asked to fill Suicidal Ideation Questionnaire-Junior (SIQ-JR) and Columbia Suicide Severity Rating Scale (C-SSRS), and take an interview with a trained social worker. |
| [23] | 588 | Army veterans | Subjects regularly complete suicide assessment questionnaire in audio, and submit the recording via an Android app. The questionnaires contains open-ended questions, where the subjects can express their views about life meaning and other suicide related topics. |
| [24] | 54 | military personnel | The subjects were interviewed for 10 minutes to 1 hour by 5 therapists specialized in suicide risk assessment, where the subjects answered Beck Scale for Suicidal Ideation (BSSI) and the Suicide Attempt Self-Injury Interview (SASII) questionnaires. The interview conversations were recorded using microphone in controlled environment. |
| [25] | 20 | Psychiatric hospital patients with suicidal history | Collected interview video recording of suicidal patients in hospital, and remote patient monitoring video using smartphones app when these patients were discharged from the hospital. |
| [26] | 43 | Suicidal patients | Following suicide patients discharge from hospital, they were asked to install android app that records their daily phone conversations. A total of 4,078 calls over 402 hours were collected. |
| [27] | 246 | Suicidal patients | Collected audio recordings of suicidal patients reading set of selected sentences, such as "my life has meaning". |
| [28] | 379 | 126 mentally ill patients, and 130 suicidal patients | Collected the video recording of every patients when answering interview questions for 8 minutes. |
| [29] | 90 | Suicidal social media users | Collected social media videos that were shared by suicidal patients. And only the videos where the subjects were talking directly to the camera is selected. |
| [30] | 40 | Suicidal students | Extracted photos of suicidal students from their high school yearbook in the past. |
| [31] | 379 | 126 mentally ill patients, and 130 suicidal patients | Collected the video recording of every patient when answering interview questions for 8 minutes. |
| [32] | 253 | Suicidal patients | Recorded the subjects' responses to standardized suicide interviews designed to harvest thought markers. |
| [13] | 60 | teenagers between the ages of 12 and 17 | Collected the video recording of subjects when answering 5 open-ended questions related to suicide ideation. The recordings were collected in a controlled examination room using a tabletop microphone. |
| [33] | 22 | Suicidal patients | Collected the audio of suicidal patients when interviewed by therapists. |
| [34] | 23 | Suicidal patients | Collected the audio recording of each patient during the interview with the therapist, as well as when the patient is reading a pre-selected rainbow passage from a book. |
| [14] | 20 | Suicidal patients | Collected the audio recording of suicidal patients during real-life situations, the recording was filtered and only high-quality recordings were considered. |
| [35] | 30 | Suicidal and depressed patients | Collected the audio recording of each patient during the interview with the therapist, as well as when the patient is reading a pre-selected rainbow passage from a book. |
| [36] | 10 | Suicidal and depressed patients | Collected the audio recording of the subject's speech when answering interview questions in a controlled environment. |
| [37] | 30 | Suicidal and depressed patients | Collected the audio recording of each patient during the interview with the therapist, as well as when the patient is reading a pre-selected rainbow passage from a book. |
| [38] | 30 | Suicidal and depressed patients | Collected the audio recording of the subject's speech when answering interview questions in a controlled environment. |
| [39] | 30 | Suicidal and depressed patients | Collected the audio recording of the patient reading a pre-selected rainbow passage from a book. The rainbow passage is selected to includes all the sounds of spoken English, and phonetically balanced. |
| [40][41] | 23 | Suicidal and depressed patients | The patients were recorded during the interview in a soundproof acoustically ideal room, where the patients were asked to take Hamilton Depression Rating Scale (HAMD) and Beck Depression Inventory (BDI-II). Following that, the patients were asked to read a rainbow passage. |
| [42] | 20 | Suicidal and depressed patients | One is a speech sample recorded from a clinical interview with a therapist and another is a speech sample recorded from a text–reading session. |

**Table 2 Audio-visual datasets for mental health research.**

| Dataset | Description | Language | Access |
|---|---|---|---|
| DIAC-WOZ [43] | Contains clinical interviews designed to support the diagnosis of psychological distress conditions such as anxiety, depression, and post-traumatic stress disorder. Data collected include audio and video recordings and extensive questionnaire responses; this part of the corpus includes data from the Wizard-of-Oz interviews, conducted by an animated virtual interviewer called Ellie, controlled by a human interviewer in another room. | English | Private |
| AVDLC [44] | Audio-visual depressive language corpus (AViD-Corpus), includes 340 video clips of subjects performing a Human-Computer Interaction task while being recorded by a webcam and a microphone. | German, English | Public |
| AVEC 2014 [45] | The Audio/Visual Emotion Challenge 2014 (AVEC) is a subset of AVDLC dataset. | German, English | Public |

### 3. Suicidal audio markers
Audio markers are classified as prosodic features, source features, formant features, and spectral features.

### 3.1. Source features
Source features are the main influencers of the changes in voice quality, changes in vocal fold vibration, loudness, vocal tract shape and pitch variation.

#### 3.1.1. Source features and suicidal
Source features measure the quality of the voice production process and observe the changes as the airflow from the lungs passes through the glottis. Table 3 lists the most used source features in automatic suicide assessment. When the patient is in a suicidal state, the laryngeal control is affected, such as harshness, breathiness and creakiness, hence this change is reflected in the source feature [46]. The source features are either extracted by measuring the changes of glottal characteristics or changes in vocal fold movements and subsequently the change in voice quality features. The majority of source features are computed by analyzing the time-series data of the glottal flow signal [47], [48]. Nonetheless, it is challenging to determine the critical time points because of the non-uniform vocal fold activity, and signal noise [49]. Source features widely used in suicidal speech analysis include the jitter, which represents the cycle-to-cycle fluctuations in glottal pulse timing during a suicidal speech; shimmer, which represents the cycle-to-cycle fluctuations in glottal pulse amplitude in voiced regions; and harmonic-to-noise ratio (HNR), which represent the ratio of harmonics to inharmonic. These source features have been proven to be correlated with suicidality [14][50], that is because they are directly related to vocal fold vibration, which influences the vocal fold tension and subglottal pressure [51].

**Table 3 Source feature used in automatic suicide assessment.**

| Feature | Description | Study | Significant test |
|---|---|---|---|

| Jitter | Deviations in individual consecutive f0 period lengths, which indicates irregular closure and asymmetric vocal-fold vibrations | [20] [23] [26][14] | Control: 0.0165 ± 0.002 ($n = 10$)<br>Suicide: 0.0217 ± 0.005 ($n = 10$)<br>$p \leq 0.05$ (t-test) |
|---|---|---|---|
| Shimmer | The difference in the peak amplitudes of consecutive f0 periods, indicates irregularities in voice intensity. | [20] [23][14] [26] | $p \leq 0.05$ |
| QOQ | The ratio of the vocal folds' opening time. Functional dysphonias often reduce QOQ range. Speaking loudly requires more effort with a low QOQ and sounds more stalled. | [21] [23] [22] [13] | Control: 0.31 ± 0.13 ($n = 8$)<br>Suicide: 0.42 ± 0.2 ($n = 8$)<br>$p \leq 0.002$ (t-test) |
| NAQ | The ratio between peak-to-peak pulse amplitude and the negative peak of the differentiated flow glottogram and normalized with respect to the period time. It can be an estimate of glottal adduction. | [21] [23][22] [13] | Control: 0.09 ± 0.04 ($n = 8$)<br>suicide: 0.12 ± 0.05 ($n = 8$)<br>$p \leq 0.002$ (t-test) |
| Spectral slope | Measures the glottal flow waveform needed for the reconstruction of the excitation signal from the given speech signal by glottal inverse filtering. [14] | [26] [14] [13] [35] | Control: −83.3 ± 5.46 ($n = 10$)<br>Suicide: −75.56 ± 8.53 ($n = 10$)<br>$p \leq 0.05$ (t-test) |
| Vocal tract | Measures the change in vocal tract dynamics and constrain articulatory movement | [37] [42] | N/A |

### 3.1.2 Source features analysis

Due to the difficulties of accurately extracting source features from speech signals, most of the works in automatic suicide detection fail to establish a strong correlation between source features only and suicide, therefore the source features are usually combined with features from other audio categories. Another influencing factor when extracting source features such as shimmer, jitter and HNR is the speech type from which these features were extracted, such as continuous speaking or held vowels context. Generally, the held vowels make the extraction of source features much easier, but they are prone to an error related to individual-related differences in sound pressure levels, which could lead to inaccuracy in feature analysis. Whereas analyzing continuous speech to extract source features is more challenging than extracting them from held vowels, as it is extremely difficult to automatically determine voiced segments in the studied utterance [52]. Identification of speech sections that contain significant source features is an active research direction in the field of suicidal speech analysis [52]. Chakravarthula et al. [20] analyzed source acoustic features in military couples' conversations and tried to deduce suicidal markers from these acoustic markers. Specifically, they have combined voice quality features (jitter, shimmer) with features from other acoustic categories, mainly prosodic features, and spectral features. They fused these acoustic features with lexical features Linguistic Inquiry and Word Count (LIWC), and behavioral features and applied a support vector machine (SVM) to classify the subjects into various classes (risk vs no-risk classes, degree of risk classes, non-severe vs severe risk classes). Scherer et al. [13] analyzed speech properties of prosodic as well as source features that were extracted from the dyadic interview corpus of both suicidal and non-suicidal adolescents. They have found various statistically significant differences between the speech properties of suicidal and healthy subjects. The source and voice quality features exhibit the strongest differences between the two subject groups. Specifically, NAQ, QOQ and peak features are strongly associated with voice qualities on the tense dimension to breathy, which show that suicidal subjects' voices are breathier than the voice of healthy people. Venek et al. [21] [7] fused source features (NAQ, QOQ) with prosodic features and formant features to analyse suicidal speech. They applied hierarchical classifier (SVM then AdaBoostM1) on speech of 60 audio-recorded dyadic clinician-patient interviews of 30 suicidal patients. They found that there were 37 statistically significant features that could distinguish suicidal from non-suicidal subjects, that including: Speech time, Pause time, Personal Pron, 1st person Pron, Impersonal Pron, Past Tense, Negation, Positive emotion, Negative emotion, Tentative, Death, Nonfluencie, Assent, NAQ and QOQ.

### 3.2. Prosodic features

Prosodic features measure longitudinal variations in a speaker's rhythm, stress level, and intonation of speech. Some of the prominent examples of prosodic features are the speaking rate, the speaker's pitch and loudness level, and the speaker's energy dynamics. Among these prosodic features, the energy, and the fundamental frequency (also known as f0, e.g. the rate of vocal fold vibration) and are the most commonly used prosodic features, as they are direct indicators of pitch and loudness.

### 3.2.1. Prosodic features and suicidal

Assessing depressed and suicidal patients based on prosodic features has been investigated in early paralinguistic studies. Depressed and suicidal patients showed prosodic speech abnormalities such as reduced pitch range, articulation errors, slower speaking rate and reduced pitch. One of the earliest efforts in this regard is the work of Dr. S. Silverman. After analyzing recorded psychiatry sessions of suicidal patients, he discovered that in pre-suicidal phase the patients' speech can be distinguished through noticeable changes in its quality [14]. Silverman noticed that speech production mechanisms of pre-suicidal patients have altered the acoustic features of speech in measurable ways [53]. Table 4 lists the most used prosodic features in automatic suicide assessment.

**Table 4 Prosodic feature used in automatic suicide assessment.**

| Feature | Description | Study | Significant test |
|---|---|---|---|
| f0 | Fundamental frequency: lowest frequency of the speech signal, perceived as pitch (mean, median). | [23][22][26] [13][33] | Control: 220.82 Suicide: 150.62 $p \leq 0.001$ [23] |
| f0 variability | Measures of dispersion of f0 (variance, standard deviation). | [21][23][22][26] [13][33] | N/A |
| f0 range | Difference between the lowest and highest f0 values. | [23][22][26] [13] [33] | N/A |
| Intensity mean | Defined as the acoustic intensity (i.e., power carried by sound per unit area in a direction perpendicular to that area) in decibels relative to a reference value, perceived as loudness. | [20] | N/A |
| Intensity variability | Measures of dispersion of intensity (variance, standard deviation). | [20] | N/A |

| Energy | Measured as the mean-squared central difference across frames and may correlate with motor coordination | [23] [24][26] [13] [39] | Control: 2.8806 (n=504) Suicide: 2.3614 (n=84) $p \leq 0.001$ [23] |
|---|---|---|---|
| Maximum phonation time | The mean of three attempts of the following measure is taken: the maximum time during which phonation of a vowel (usually lal) is sustained as long as possible with an upright position, deep breath, and a comfortable pitch and loudness | [23] | N/A |
| Speech rate | Number of speech utterances per second over the duration of the speech sample (including pauses). | [21] | N/A |
| Time talking | Sum of the duration of all speech segments. | [21] | N/A |
| Pause duration mean | Mean duration of pause length. | [21] | N/A |
| Pause variability | Measures of dispersion of pause duration (variance, standard deviation). | [21] | N/A |
| Pause rate | Total length of pauses divided by the total length of speech (including pauses). | [21] | N/A |
| Pauses total | Total duration of pauses. | [21] | N/A |
| Pitch | The pitch level | [20] [23][24] [26] [29] | N/A |
| PSP | Measure is derived by fitting a parabolic function to the lower frequencies in the glottal flow spectrum. | [21] [22] | Control : 0.36 (n= 504) Suicide : 0.50 (0.09) (n=84) $p \leq 0.05$ [22] |
| MDQ | The glottal closure instants (GCI) and the dispersion of peaks in relation to the GCI position is averaged across different frequency bands and then normalized to the local glottal period which outputs the MDQ parameter . | [22] | N/A |

**3.2.2 Prosodic features analysis**

Among prosodic features F0 is the most used one. France et al. [33] studied the prosodic feature (f0), along with other acoustic features extracted from two recording samples, male subjects sample and female subjects samples. They found that formant and spectral features are the best discriminators between male and female samples, and that prosodic feature is the most discriminator within male subjects. Belouali et al. [23] analysed prosodic features to study the statistical significance of the correlation of these features in army veterans' daily conversation and suicidality. A fused set of 15 acoustic features extracted from the speech were measured by the ensemble feature selection. Random Forest (RF) classifier was applied to determine suicidal group, which yielded 86% sensitivity and 70% specificity. Gideon et al. [26] extracted eGeMAPS features that include prosodic features along with emotional features obtained from the speech from recordings of natural phone conversations to detect suicidal callers. They could extract low-level descriptors (LLDs) for frequency, amplitude, spectral and energy, the parameters in every speech section, which yielded 23 values per frame.

### 3.3. Formant features

Formant features are also known as filter features, which measure the resonant characteristic of the nasal and vocal tracts that filter the source coming from the vocal folds. The shape of the nasal and vocal tracks are used to reduce some frequencies and increase other frequencies.

### 3.3.1. Formant features and suicidal

As formant features include vocal tract and acoustic resonance information, hence these features can capture the changes in the suicidal patient's speech changes, if these changes are reflected in mucus secretion and muscle tension. Early studies reported that suicidal speech can be associated with increase in formant frequency and decrease in its bandwidth. In addition to that, suicidal and depressed patients have different, as suicidal patients had shifts in power from lower to higher frequency compared to depressed patients [33]. Table 5 lists the most used formant features in automatic suicide assessment.

**Table 5 Formant feature used in automatic suicide assessment.**

| Feature | Description | Study | Significant test |
|---|---|---|---|
| F1 | The first peak in the spectrum (especially of voiced utterances such as vowels) results from a resonance of the human vocal tract. | [22][26][27] [29] [33] | Control : 0.1893 (n= 504) Suicide : 0.1805 (n=84) $p \leq 0.001$ [22] |
| F2 | The second peak in the spectrum (especially of voiced utterances such as vowels) results from a resonance of the human vocal tract. | [26] [29] [33] | N/A |
| F1 variability | Measures of dispersion of F1 (variance, standard deviation). | [21] [22][26] [27] [29] [33] | N/A |
| F2 variability | Measures of dispersion of F2 (variance, standard deviation). | [21][26] [29] [33] | N/A |
| Line spectral features | Measure the linear correlation between spectral frequencies | [20] | N/A |
| Spectral flex | Measures the flex ratio in spectral. | [26] [29] | N/A |
| Spectral stationarity | Measures the range of the | [13] | N/A |

| | prosodic inventory used over utterances and the monotonicity of the speech | | |

### 3.3.2. Formant features analysis

Many studies have found associations between formant feature changes and suicidal state. For instance, the early study by France et al. [33] showed that the increase in formant features, specifically the formant frequencies and the decrease in formant bandwidth are higher in suicidal patients. Stasak et al [27] extracted the formant F1 features, in addition to the disfluency and voice quality features from the speech of 226 psychiatric inpatients with a recorded suicidal behavior. They manually annotated the recorded speeches and converted them to low-dimensional vectors, which helped to identify suicidal patients with more than 73% accuracy. Shah et al. [29] extracted the F0, F1 and F2, spectral flux, loudness and average temporal interval of voiced, as well as silent speech sections. Every feature was averaged over the total duration of the video yielding scalar features. Variance in pitch was measured as the standard deviation in the F0 values in voiced sections.

### 3.4. Spectral features

Spectral features measure the spectrum of the subject's speech; which represents the frequency distribution of the speech signal at a given time. Some of the widely used spectral features for suicidal speech analysis include Mel Frequency Cepstral Features (MFCCs) and Power Spectral Density (PSD). Many previous studies have proven the relationship between spectral features, such as energy shift that is measured by PSD, and suicidal speech [42]. Akkaralaertsest et al. [35] extracted and fused spectral features, including the Glottal Spectral Slope (GSS) and MFCC from the voiced section of the speech sample database. Following that, they combined these spectral features with other acoustic features, mainly source and formant features to a classification model. Their results prove that by combining features from the speech production system (source features), along with spectral features can largely increase the classification accuracy. Keskinpala et al [39] extracted and analyzed the spectral features of male and female speech samples from high-risk suicidal patients, specifically, they analyzed MFCC and energy in frequency bands. Their results suggest that mel-cepstral coefficients and energy in frequency band features are highly effective to distinguish between depressed and suicidal patients. Anunvrapong et al. [36] leveraged MFCC and Delta-MFCC (ΔMFCC) spectral features. They extracted and analyzed ΔMFCCs which is the sixteen consecutive MFCCs, then trained a classifier using various speech datasets of depressed and suicidal patients. Their proposed classifier can distinguish between suicidal and depression patients with 95% accuracy. Ozdas et al. [38] extracted MFCC features by dividing the voiced speech segments and measuring the logarithm of the discrete Fourier transform (DFT) of every speech segment. Every log spectrum is further filtered with 16 triangular filters. Using only the MFCC features, they could reach 80% accuracy in classifying near-term suicidal patients and non-depressed subjects. Similarly, Wahidah et al. [40] used PSD and MFCC features and achieved 79% accuracy in classifying suicidal patients. Table 6 lists the most used spectral features in automatic suicide assessment.

**Table 6 Spectral feature used in automatic suicide assessment.**

| Feature | Description | Study | Significance test |
|---|---|---|---|
| MFCC | The coefficients derived by computing a spectrum of the log-magnitude Mel-spectrum of the audio segment. The lower coefficients represent the vocal | [20][23] [26] [35] [38] [39] [40] [41] [54][55] | N/A |

| | tract filter and the higher coefficients represent periodic vocal fold sources. | | |
|---|---|---|---|
| ΔMFCC | Measures rapid temporal information captured in the MFCC extraction | [36][23] | Control: 0.1893 (n= 504) Suicide: 0.1805 (n=84) $p \leq 0.001$ [23] |
| PSD | describes how the power of your voice is distributed over frequency | [34] [39] [40] [41] [42] | N/A |
| GSS | Measures the periodogram of the voiced segments. | [35] | N/A |
| PS | The feature is essentially an effective correlate of the spectral slope of the speech signal. | [15] [22] | Control : -0.20 (n= 504) Suicide : -0.24 (n=84) $p \leq 0.05$ [22] |

## 4. Suicidal visual markers

Visual markers are classified as facial features; eye movement features or posture features.

### 4.1. Facial features

A growing literature in mental disorder detection has proven that facial appearance and facial expression can carry significant non-verbal cue that can be further interpreted to assess various mental disorders such as depression [8], bipolar [56] and social anxiety [57]. Given that mental disorders are driving factors of suicide, the facial expression is an important non-verbal suicidal cue for distinguishing suicidal patients [58]. Kleiman et al. [30] analyzed the facial feature extracted from photos of 40 people who committed suicide. After performing a t-test, they found that accuracy in discerning whether a subject had committed suicide was significantly higher than chance guessing by participants. Laksana et al. [28] extracted various visual features including frowning, smiling, head movement and eyebrow-raising behaviors. They investigated the occurrence and the frequency of these behaviors. The results suggest that facial expressions such as smiling behaviors are more statistically correlated with suicide than other visual features such as eye movements and head movements. Eigbe et al. [31] studied the smile dynamics (e.g. genuine vs fake smiles) of three groups: people with suicide ideation, people with depression and healthy control subjects. They found that suicidal subjects had the shortest smiling time duration, and the healthy control subject had the longest smiling time. Moreover, the percentage of genuine smiles in suicidal subjects was the lowest, and the healthy control subjects also had a higher percentage of speaking smiles and the longest laughing duration compared to depressed and suicidal groups.

### 4.2. Eye movement features

Eye movement features have been proven to have a strong correlation with suicidal intent, especially when the suicidal patients are intimidated by a suicide-related question that they often try to avoid answering. In such circumstances, the patients tend to avoid eye contact with the interviewer, with frequent gazing down behaviors. Eigbe et al. [31] studied the gaze aversion (e.g. looking down) of three groups: people with suicide ideation, people with depression and healthy control subjects. They found that the depressed group had significantly less gaze down count than the suicidal group and healthy control group. Additionally, suicide group subjects have higher gazing down, they also found that suicidal patients spent a higher frequency of gazing down than control patients, and their gazes time were longer than other groups as well. Shah et al. [29] investigated the eye gaze movement and concluded that the frequent shift in eye gaze aversion and eye gaze represents a vital behavioral indicator, as they reflect the subjects' social avoidance, which is strongly correlated with suicidal ideation.

### 4.3. Posture features

Although posture features extracted from body movement have been proven to be a strong indicator of depression [8], very few works have studied the relationship between body movement and suicidal behaviors. Galatzer-Levy et al. [25] used OpenFace to extract the angle of the head's yaw (horizontal movements) and head's pitch (vertical movement) from a video recording of suicidal patients' interviews. After analyzing the head movement along with vocal and facial features, and their correlation with suicidality, their results indicate that head yaw and head pitch variability have significant negative linear relationships with suicidality and that low levels of head movement is strongly correlated with suicide severity. Laksana et al. [28] studied head movements that are interpreted as anxious expressions, such as fidgeting, and looking around the room, and their relationship with suicide ideation. They observed that suicidal patients often exhibit anxious expressions and their high head velocity is higher than healthy control subjects that remain relatively stable during the interview. Table 7 summarizes the most used visual feature for automatic suicide detection.

**Table 7 Visual suicidal markers**

| Class | Feature | Studies |
|---|---|---|
| Facial | Smile Intensity | [28][8][59] |
|  | Smile duration | [8][59] |
|  | Duchenne Smile Percentage | [28] |
|  | Sharpness of Smile Onset/Offset | [28] |
|  | Frowning Behavior | [28] |
|  | Eyebrow Raises | [28] |
|  | Facial expression (Action unit) | [25][60] |
|  | Smile | [31] |
|  | frequent changes in facial expression | [29] |
|  | Mouth nose distance | [61] |
|  | Mouth landmark/ mouth centroid | [62] |
|  | Facial landmark velocity/acceleration | [63] |
|  | Polynomial fitting | [63] |
| Eyes | shift in eye gaze | [29] |
|  | eye gaze aversion | [29][64] |
|  | gazing down behavior | [31] |
|  | pupil location | [29] |
|  | gaze angle | [29] |
|  | frequently shifting gaze | [29] |
|  | eye pupil movement | [61] |
|  | blinking frequency | [61] |
|  | Eyebrows and mouth corners movement | [61] |
|  | Distance between eyebrows | [61] |
|  | Distance upper-lower eyelid | [62] |
|  | Average vertical gaze | [8] [59] |
|  | Vertical eyelid movement | [64] |
| Posture | Head Motion Velocity | [28] [64] |
|  | head's pitch (up and down movement) | [25] [64] |

|  | Head yaw (side to side movement) | [25] [64] |
|  | STIP features (upper body movement) | [65][66] |

In Table 8, we summarize the above-reviewed research in terms of data source, used acoustic and visual features, analysis method, and targeted suicide type. Descriptive statistical methods tried to gain insight into the different audiovisual features and the presence of audiovisual markers in different groups (e.g. suicidal group vs healthy control group). Machine learning methods on the other hand are more focused on the automatic prediction of suicidal patterns in the input features in raw audiovisual data, even in the absence of context (e.g. patient group). Although we cannot rely on statistical analysis alone to draw general conclusions regarding audiovisual features and their relevance to specific suicidal markers, statistical insights have the potential to boost the training process of machine learning models by reducing the training time and increasing predictive performance. For example, this can be accomplished by leveraging the predictions of multiple different models trained on independent datasets from different clinical environments and aggregating them under a single modeling framework.

**Table 8 Audio-visual suicide assessment scheme**

| Ref | Data source | Acoustic features | Visual features | Analysis method | Targeted suicide type |
|---|---|---|---|---|---|
| [20] | Conversation | MFCCs<br>line spectral frequencies<br>jitter<br>shimmer | N/A | SVM | Passive suicide ideation<br><br>Suicidal action |
| [21][6][22] | Interview | NAQ<br>QOQ<br>PSP<br>MDQ<br>PS<br>Formants (F1, F2) | N/A | SVM<br><br>AdaBoostM1 | Passive suicide ideation |
| [23] | Interview | MFCC, Energy, Amplitude, F0, jitter, shimmer, amplitude perturbation quotient, pitch perturbation quotient, logarithmic energy, QOQ and NAQ | N/A | logistic regression (LR);<br><br>random forest (RF);<br><br>SVM<br><br>XGBoost (XGB);<br><br>k-nearest neighbors (KNN);<br><br>deep neural network (DNN) | Passive suicide ideation<br><br>Active suicide ideation<br><br>Suicidal action |
| [24] | Interview | pitch<br><br>energy<br><br>spectral feature | N/A | Lyapunov coefficient | Active suicide ideation |
| | | voice quality | | | Suicidal action |
| [25] | Interview | speech prevalence | The angle of the head's pitch (up and down movement) yaw (side to side movement) Facial Action Coding Scheme (FACS) | multiple linear regression | Suicidal action |
| [26] | Conversation EMA | eGeMAPS features | Emotions | DNN (emotion recognition) | Active suicide ideation Suicidal action |
| [27] | Reading task | F1 Disfluency GRBASI voice quality | N/A | SVM KNN | Active suicide ideation Suicidal action |
| [28] | Interview | N/A | Smiling, Frowning Behavior, Eyebrow Raises, Head Motion Velocity | SVM Random Forest, Multinomial Naıve Bayes. | Active suicide ideation Suicidal action |
| [29] | Conversation | F0, F1, F2, loudness, spectral flux, pitch variations | shift in eye gaze eye gaze aversion pupil location gaze angle frequently shifting gaze | Multimodal predictive modelling | Active suicide ideation |
| [30] | Images | N/A | Facial features | Signal detection theory | Suicidal action |

| Ref | Speech type | Acoustic features | Visual features | Method | Suicide category |
|---|---|---|---|---|---|
| [31] | Interview | N/A | Smile<br>Gazing down behavior | Wilcoxon Rank Sum test | Passive suicide ideation<br>Active suicide ideation<br>Suicidal action |
| [13] | Interview | Energy<br>f0<br>Peak slope<br>Spectral stationarity | N/A | HMM<br>SVM | Passive suicide ideation<br>Active suicide ideation<br>Suicidal action |
| [33] | Interview | f0<br>Amplitude modulation<br>Formants<br>Power distribution | N/A | Autoregressive model | Passive suicide ideation |
| [34] | Read speech<br>interview | Power spectral densities<br>peak power, peak location,<br>PSD | N/A | Analysis of variance (ANOVA) | Passive suicide ideation |
| [14] | Natural speech | Jitter<br>Glottal flow spectrum | N/A | Maximum Likelihood Classifier | Suicidal action |
| [35] | Interview<br>Read speech | Glottal Spectral Slope<br>MFCC | N/A | PCA<br>Least Squares (LS) | Passive suicide ideation |
| [36] | interview | MFCC | N/A | ML and LMS | Passive suicide ideation |
| [37] | Interview | Vocal-tract articulation | N/A | GMM | Passive suicide ideation |
| | Read speech | | | | |
| [38] | Interview | MFCC | N/A | GMM | Passive suicide ideation |
| [39] | Interview Read speech | MFCC, Energy in frequency bands (power spectral density). | N/A | Unimodal Gaussian modelling | Passive suicide ideation |
| [40] [41] | Interview Read speech | PSD MFCC | N/A | Multiple linear regressions | Passive suicide ideation |
| [54] | Interview Read speech | MFCC | N/A | GMM | Passive suicide ideation |
| [42] | Interview Read speech | PSD Vocal-tract articulation | N/A | GMM | Passive suicide ideation |

## 5. Artificial intelligent suicide detection models

To detect suicidal signs from audiovisual data, most of the existing works applied either statistical analysis to describe the uniqueness of suicidal patients compared to healthy subjects, or machine learning classifiers to infer the commonalities among suicidal patients. Statistical analysis is usually performed to infer meta-knowledge from the extracted audiovisual features. For instance, Doval et al. [24] modeled the speech feature stream as the observed variable of a nonlinear dynamical system, then they calculated the largest Lyapunov coefficient and correlation dimension of the speech series. Gideon et al. [26] investigated the statistical relationship between suicidal ideation and emotion estimated from speech. Specifically, they computed the within-subject standard deviation of each emotion to gauge each emotion's variability. Similarly, Shah et al. [29] computed GNorm distance to identify the statistical significance between suicidal and non-suicidal groups. ML classification methods are used to categorize the subjects with similar audiovisual features and are expected to distinguish suicidal patients from healthy subjects. Surprisingly, deep-learning models are rarely used in the literature on suicide detection from audiovisual data, despite their popularity in similar tasks such as depression detection [67], as deep-learning models have been proven to achieve good results when applied to audiovisual data. Overall, automatic suicide assessment can reduce diagnosis cost, offer instantaneous detection and avoids human biases.

### 5.1. Linear classifiers

SVM are supervised learning models with associated learning algorithms that analyze data for classification tasks, as well as regression analysis. Although SVM is most used for linear classification tasks, but it can also deal with non-linear classification using what is called the kernel trick, implicitly mapping their inputs into high-dimensional feature spaces [68]. SVM is by far the most used for classifying suicidal patients [13], [20]–[23], [27], [28], [31][55], this is due to the fact that SVM requires just small set of training data, which made it desirable for the aforementioned works. Chakravarthula et al. [20] leveraged SVM classifier to categorize the subjects into various classes (risk vs no-risk, degree of risk, non-severe vs severe risk). They applied sample weighting to mitigate class imbalance and tuned hyperparameters such as feature normalization scheme. Scherer et al. [9] compared SVM with HMM to classify interview recorded audio into suicidal/Non-suicidal. They reported that SVM can achieve accuracy of up to 75%, but HMM could achieve up to 81.25% classification accuracy. Similarly, Stasak et al. [27] used an automatic 2-class classification using SVM and non-linear cosine k-nearest neighbor (KNN) algorithm. While some other works [37], [38], [42] have applied GMM) to model the suicidal patients' class distributions of the extracted audiovisual features.

### 5.2. Boosting classifiers

Ensemble methods leverage more than one weak learner algorithm to improve generalizability and overcome the drawback of a single weak estimator. Boosting is an ensemble technique primarily used to reduce bias that converts weak learners to strong ones. Boosting classifier gives relatively high accuracy when combined more than weak classifiers such as SVM, for suicide/non suicide groups classification. For example, Venek et al. [21][22] proposed a hierarchical classification method with two layers. In the first layer, the suicidal patients and the non-suicidal patients are distinguished using SVM classifier. In the second layer, the suicidal repeaters and the non-repeaters are classified using the ensemble algorithm AdaBoostM1. Using only the patients' features, the classification of suicidal vs. non- suicidal patients achieve an accuracy of 85%. However, by adding the clinician's features the accuracy was improved to 86.7%.

### 5.3. Decision trees and random forests

Decision trees are non-parametric supervised learning methods that can be used for classification. The rationale behind using tree structure is to create a model that predicts the value of a target variable by learning simple decision rules deduced from the data features. A tree can be viewed as a piecewise constant approximation. A random forest is an ensemble method that combines many decision trees at training time. In the context of classification, the output of the random forest is the class selected by most trees. Belouali et al. [23] applied RF classifier on a fused set of 15 acoustic features extracted from the speech to determine the suicidal group, which yielded 86% sensitivity, 70% specificity. However, Laksana et al. [28] reported that SVM performed better than RF when applied to a single visual feature (smiling).

### 5.4. Deep neural networks

Deep neural networks refer to a class of supervised/unsupervised algorithms that are modeled to imitate the human brain, that are designed to recognize patterns. In the context of a classification task, DNN reads the data through the input layer of multiple perceptrons, passes it through stacked hidden layers, and finally the output layer determines the class of the input. DNN performs better than conventional classifiers when the dataset is relatively large. DNN have been reported to achieve high performance when dealing with audiovisual data. Belouali et al. [23] compared six different supervised classification algorithms on audiovisual to classify suicidal/non suicidal patients, namely DNN, logistic regression, RF, SVM, XGB and KNN. When applied to acoustic features only, DNN yielded the highest specificity 70% compared to other classifiers, but the lowest specificity 50% when applied to linguistic features only, which confirms that DNN are more suitable for audiovisual data.

In Table 9, we compare the performance of different ML classifiers in the context of suicidal assessment using audiovisual.

**Table 9 ML classifiers for suicidal classification**

| Classifier | Advantage | Drawbacks | Best reported accuracy |
|---|---|---|---|
| SVM | Effective in high dimensional spaces. Can tolerate the lack of data, and even if where the number of features is greater than the number of samples. Memory efficient. | Do not take previous observations into account and are trained on the median and standard deviations of the features over the single utterances. [9] Do not directly provide probability estimates. Prone to overfitting | 75% accuracy [9] 60.32 recall [20] 86.7% accuracy [21] (SVM-AdaBoostM1) |
| HMM | Can take advantage of the sequential and dynamic characteristics of the observations and classify each segment on the full frequency sampled feature vector. [9] | Memory consuming | 81.25% accuracy [9] |
| AdaBoostM1 | Mitigate the bias of single weak learner Resilient to over-fitting | Sensitive to outliers Difficult to scale up, because every estimator bases its decisions on the previous predictors. | 86.7% accuracy [21] (SVM-AdaBoostM1) |
| RF | Simple to understand and to interpret Can tolerate the lack of data The model can be validated using statistical tests. | Prone to overfitting Create biased trees if some classes dominate | 86% sensitivity [23] |
| DNN | Suitable for audiovisual data | Requires large dataset | 70% specificity [23] |

## 6. Discussion, limitations and future directions
In this section, we summarize the finding of this review, and discuss the limitation, as well as potential future directions.

### 6.1. Discussion
Recording settings: The audiovisual data recording is conducted in a controlled, semi-controlled, or free environment. In controlled settings, the recording is conducted inside an acoustically ideal environment such as an anechoic chamber with a high-quality microphone, and in some cases, the patients were explicitly instructed not to drink caffeinated drinks or alcoholic beverages for several hours before the recording to ensure the use of these did not influence their voice quality. While in semi-controlled environment, the requirement of the soundproof chamber and high-quality microphone is relaxed, therefore the recording can be done remotely, however, the subjects are still instructed to follow some guidelines, such as being in an empty room, or facing a laptop during the recording. Finally, in the free environment, the recording is conducted in any circumstances, usually using smartphones, which gives the advantage of interacting with the patients in natural environments, hence minimizing recall bias and maximizing ecological validity, and allowing the study of micro-processes that influence behavior in real-world contexts [69]. On the other hand, this poses a great challenge in processing the voice in an uncontrolled environment. Which involves extra work on noise removal and acoustic feature pre-processing, due to the low quality of smartphone microphones compared to dedicated microphones [70].

Important audiovisual features: Among all the acoustic features, MFCC and F0 is the most used feature that has been strongly correlated with the features and has consistently been observed to change with a speaker's mental state. And when MFCCs are inputted to GMM as an effective mechanism of speech analysis, this combination has been proven to be ideal for classifying either high-low levels of suicidality. Facial expression is the most effective feature that has been correlated with suicide. Most of the facial features used in the reviewed works are dynamic features, where the changes in the facial feature are observed over a period of time, such as smile intensity or eye movement, which require video recording. Therefore, we clearly observe that works that used video have generally high accuracy compared to works that use only static visual features extracted from images. Furthermore, works that used multimodal features from different categories or even within the same category yield better accuracy compared with single modality schemes, e.g., combining multiple audio features from different categories or visual cues from face and upper body for instance.

Standardized audiovisual processing framework: Preprocessing, extracting and manipulation of audiovisual features can pose many challenging tasks. However, the usage of collaborative and freely available repositories for speech and video processing frameworks can facilitate boost research in this field. For acoustic feature processing, the collaborative voice analysis repository for speech technologies (COVAREP) [71] and The Geneva minimalistic acoustic parameter set (GeMAPS) [72] are the most popular tools of acoustic feature manipulation. For the visual feature, one of the prominent tools is OpenFace library, which is capable of facial landmark detection, head pose estimation, facial action unit recognition, and eye-gaze estimation.

In Table 10, we summarize the strengths and weaknesses of automatic assessment compared to clinical assessment.

**Table 10 Advantages and disadvantages of clinical vs automatic suicide assessment**

| Method | Advantage | Disadvantage |
|---|---|---|
| Clinician assessments | Clinician experience<br>Clinicians can ask for further assessments<br>Questionnaire items are interpretable<br>Clinicians can offer treatment pathway | Relatively expensive (Costs of clinic and clinician)<br>Questionnaires often use ordinal and vague variables (eg, never, sometimes)<br>Time-consuming<br>Prone to clinician's biases (e.g. expertise, culture and race)<br>Difficult to measure and quantify complex features |
| Automatic assessment | Instantaneous detection<br>Low cost<br>Can be performed remotely and continuously.<br>Avoids human biases and single rater.<br>Can capture multimodal features (e.g. audio-visual)<br>Can process complex features using linear/nonlinear multivariate models, and discover new structure in data.<br>Allows scalability because models can be fast and automated | Requires large datasets for training and testing.<br><br>Most models have not been validated through clinical trials so far.<br><br>Models accuracy can be affected by biases in data (eg, race, gender, age, noise) |

## 6.2. Ethics and privacy

Developing AI-based suicide detection models that can be applied to real-time systems, such as video streaming and calling apps, raises serious privacy concerns from a research perspective as well as a deployment perspective. From a research point of view, the users' privacy might be jeopardized throughout the data processing stages. When dealing with publically available datasets, the problem arises when users' personal attributes can be predicted, and the identity of subjects can be revealed. In this context, many jurisdictions require certain conditions for research that can compromise users' privacy. The most common procedure is that the researchers must acquire an ethical approval or exemption from their Institutional Review Board (IRB) before the study. In addition to that, they must obtain informed consent from the participant when possible, protect, and anonymize sensitive data during all research stages. Moreover, they need to be careful when linking data across sites is necessary. Finally, when sharing their data, they need to make sure that other researchers also adhere to the same privacy guidelines [73]. Researchers can rely on public social media datasets for AI-based suicidal behavior research as long as they ensure the preservation of the confidentiality of users. The privacy concern is more challenging when applying such AI-based solutions in real-time and large-scale scenarios.

## 6.3. Limitation

The limitations of reviewed works on audio-visual data analysis for suicidal behavior detection can be summarized as follows:

Large-size dataset and population generalization: One of the main limitations of some related works is the relatively small size of the dataset used to train the proposed models, both in terms of the number of patients and recorded duration. E.g. n=20 [25], n=90 [29], n=97 [24], n=126 [28], n=124 [20]. Morever, many of the previous works have focused their studies on specific high-risk groups, and the dataset that they used to generate the results and draw the conclusion is related only to that specific group. For example, army veterans [23][24], Adolescent [22], LGBT [74]. However, acoustic variability is highly affected by linguistic information and speaker characteristics such as age, gender and cultural background. Therefore, the challenges in finding a speech-based marker for suicide require the study of a large population from different groups to tune and generalize the proposed suicide assessment tool.

Lack of public datasets: Most of the relevant public datasets, including the datasets presented in Tabel 2, are general-purpose datasets that have been collected in the context of detecting abnormal behavior that can be latent symptoms of mental illnesses. The lack of public, or even private, datasets that have been collected specially for suicidal behavior detection, is one of the main limitations in this line of research.

## 6.4. Future directions

Unlike text-based suicidal assessment that has been widely studied, suicide assessment through audio-visual data analysis is still in early development. There are many potential future research directions in this area:

- Conversational AI for suicide assessment: conversational AI agents can engage in a conversation with a user through written text or voice. Conversational AI is an important technology that has the potential to dominate the area of automatic suicide detection and intervention. As suicidal patients may feel more comfortable expressing their thoughts to AI agents than to a human. The future direction in this area is to train AI chatbots on a large data-set of suicide notes, and use this pre-trained model and customize it to the characteristics of each patient.

- Multimodal deep-learning for suicide detection: Deep-learning models have been proven to achieve good results when applied to audiovisual data. Applying deep-learning to multimodal features from different audio-visual categories or even within the same category usually yield better accuracy compared with single modality schemes. For example, conventional neural networks (CNN) are effective when dealing with images, video or audio, while recurrent neural networks (RNN) models such as LSTM and GRU are effective in handling time-series data. One prominent future direction is applying a hybrid CNN-RNN model that has been pre-trained on video interviews of confirmed suicidal patients, such a pre-trained model can capture the time-series markers of suicidal behaviors.

## 7. Conclusion

In this paper, we have reviewed the recent works that studied suicide ideation and suicidal behavior detection through audiovisual feature analysis, mainly suicidal voice/speech acoustic features analysis and suicidal visual cues. Audiovisual features have been proven to convey significant non-verbal cues that can be further interpreted to assess various mental disorders, however, the literature of audiovisual based suicide assessment is still relatively limited to a few small-scale experiments. Automatic suicide assessment is a promising research direction that is still in the early stages. There is a lack of large datasets that can be used to train machine learning and deep learning models that have been proven effective in other similar tasks. Building an AI-enabled suicide risk assessment tool that observes the patient's biomarker and audiovisual cues and learns suicide ideation patterns might be a decisive indicator that could help the clinician to reach a decisive judgment.